\documentclass{article}

 \usepackage{graphicx}
 \usepackage{wrapfig}
 \usepackage{amsmath}
 \usepackage{subcaption}
\usepackage{wrapfig}
\usepackage[ruled,vlined]{algorithm2e}
\usepackage{amsmath}
 \usepackage[preprint]{neurips_2025}

\usepackage[utf8]{inputenc} 
\usepackage[T1]{fontenc}    
\usepackage{hyperref}       
\usepackage{url}            
\usepackage{booktabs}       
\usepackage{amsfonts}       
\usepackage{nicefrac}       
\usepackage{microtype}      
\usepackage{xcolor}         

\title{Data-Dependent Smoothing for Protein Discovery with Walk-Jump Sampling}

\author{%
Srinivas Anumasa$^1$, Barath Chandran.C$^2$, Tingting Chen$^1$, Dianbo Liu$^1$ \\ 
   \textsuperscript{1}National University of Singapore  \textsuperscript{2}Indian Institute of Technology, Roorkee\\
  \texttt{srinu\_pd@nus.edu.sg} \\
  \texttt{barath\_cc@ece.iitr.ac.in} \\
}

\begin{document}

\maketitle

\begin{abstract}

Diffusion models have emerged as a powerful class of generative models by learning to iteratively reverse the noising process. Their ability to generate high-quality samples has extended beyond high-dimensional image data to other complex domains such as proteins, where data distributions are typically sparse and unevenly spread. Importantly, the sparsity itself is uneven. Empirically, we observed that while a small fraction of samples lie in dense clusters, the majority occupy regions of varying sparsity across the data space. Existing approaches largely ignore this data-dependent variability. In this work, we introduce a Data-Dependent Smoothing Walk-Jump framework that employs kernel density estimation (KDE) as a preprocessing step to estimate the noise scale $\sigma$ for each data point, followed by training a score model with these data-dependent $\sigma$ values. By incorporating local data geometry into the denoising process, our method accounts for the heterogeneous distribution of protein data. Empirical evaluations demonstrate that our approach yields consistent improvements across multiple metrics, highlighting the importance of data-aware sigma prediction for generative modeling in sparse, high-dimensional settings.
 
\end{abstract}

\section{Introduction}
\begin{wrapfigure}{r}{0.44\textwidth}
    \centering
 
    \begin{subfigure}[b]{0.20\textwidth}
        \includegraphics[width=\linewidth]{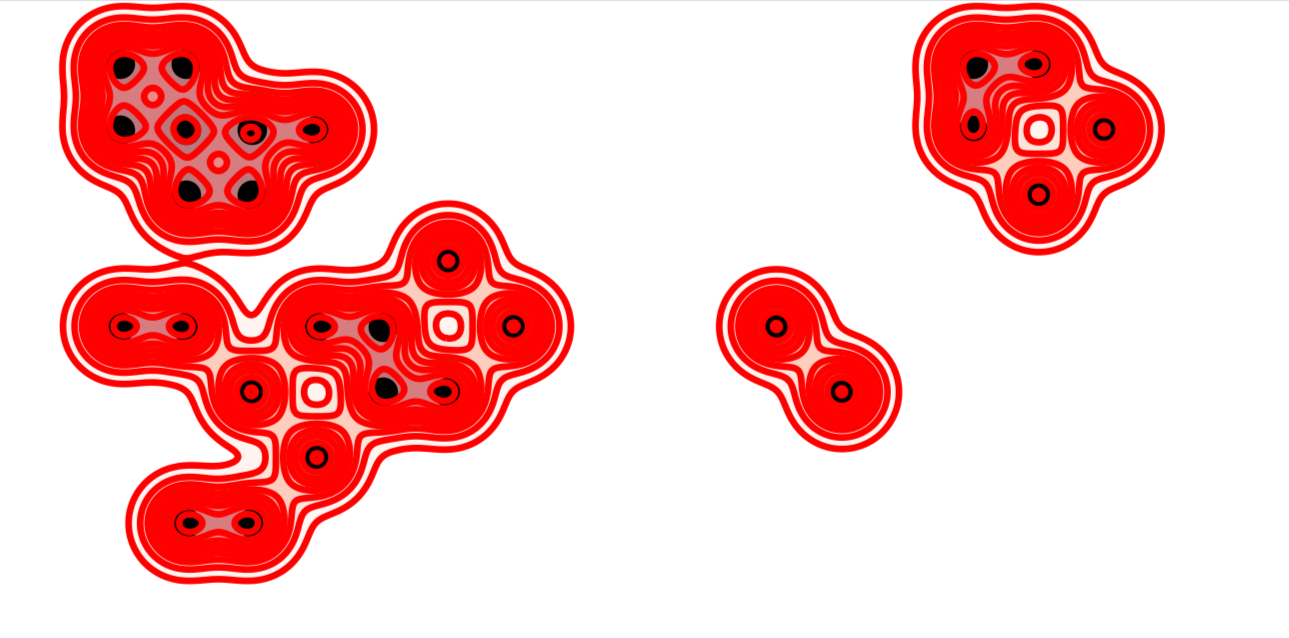}
       \caption{}
    \end{subfigure}
    \hfill
    \begin{subfigure}[b]{0.20\textwidth}
        \includegraphics[width=\linewidth]{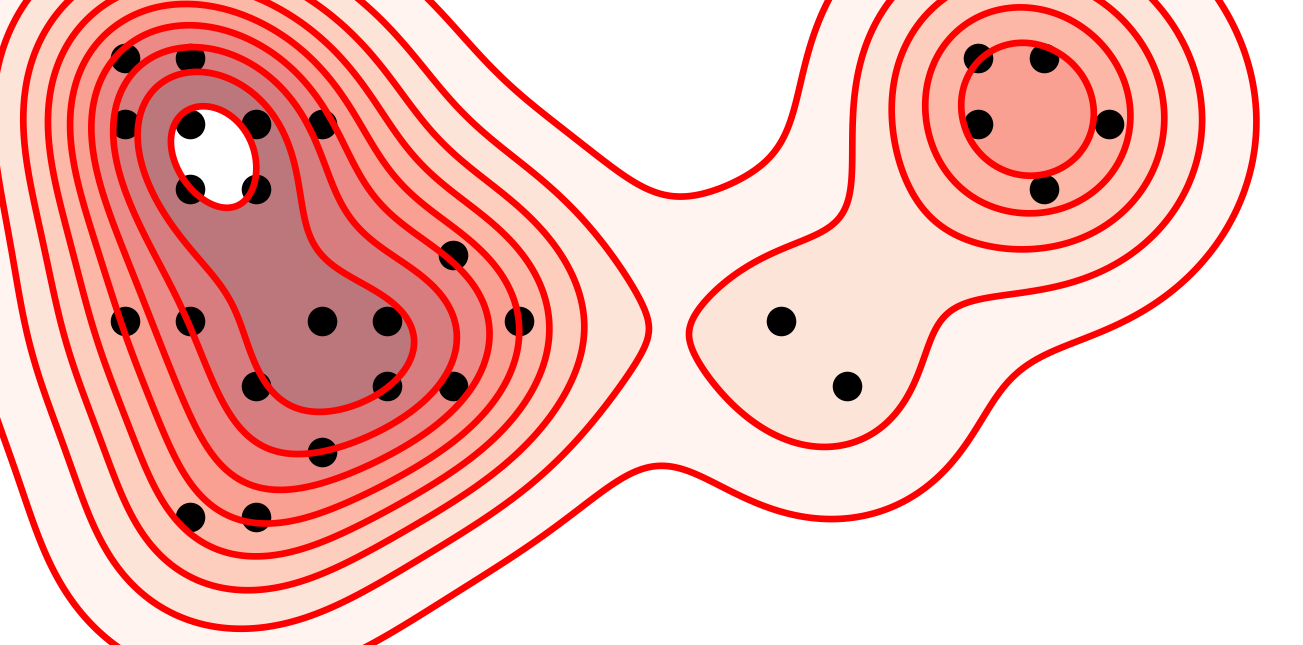}
        \caption{}
    \end{subfigure}

    \begin{subfigure}[b]{0.20\textwidth}
        \includegraphics[width=\linewidth]{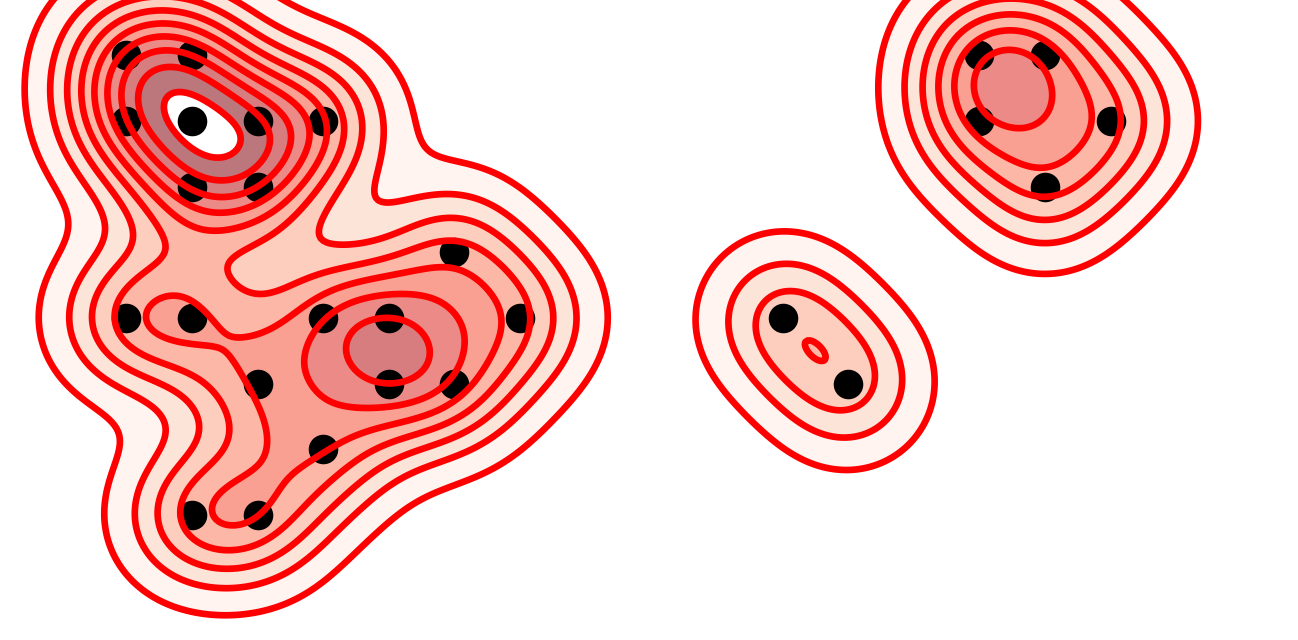}
        \caption{}
    \end{subfigure}
    \hfill
    \begin{subfigure}[b]{0.20\textwidth}
        \includegraphics[width=\linewidth]{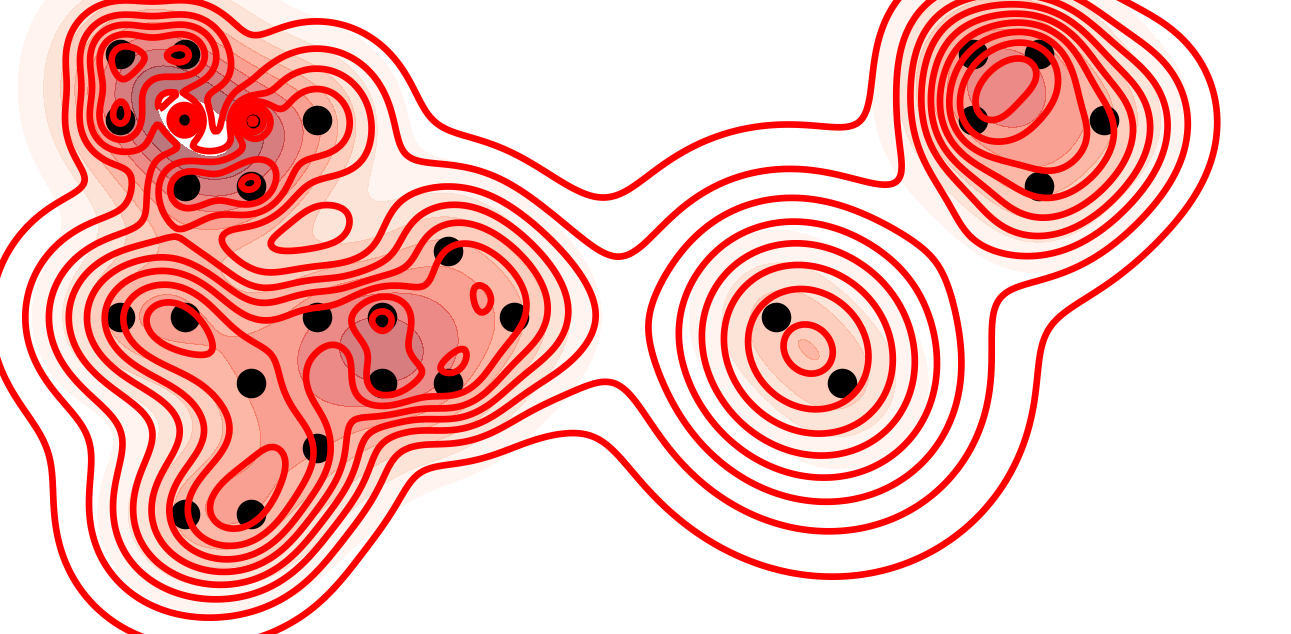}
        \caption{}
    \end{subfigure}
    
    \caption{Illustration of different contour behaviours: (a) small $\sigma=0.4$ oversharpens, 
    (b) large $\sigma=1.5$ oversmooths, 
    (c) With Potential Optimal $\sigma=0.8$
    (d) Our adaptive $\sigma$ mapping.}
    \label{fig:intro_contours}
\end{wrapfigure}
Sampling in high-dimensional spaces is widely regarded as a challenging task due to the curse of dimensionality \cite{bishop2006pattern}. Recently, diffusion models have emerged as a powerful class of generative models capable of producing high-quality samples across diverse domains, including images \cite{ho2020denoising,dhariwal2021diffusion}, videos \cite{ho2022video}, graphs \cite{jo2022score}, and biological sequences \cite{frey2023protein}. The central training objective of diffusion models is to learn the score function, the gradient of the log-probability density function (PDF) using neural networks\cite{song2020sliced}. Usually, the training relies on estimating the score of noisy distributions, obtained by convolving the data distribution with known noise distributions (typically Gaussian). The neural network thus learns to predict score values by denoising perturbed samples. At inference time, starting from pure noise, the learned denoiser iteratively refines samples through a reverse diffusion process, ultimately recovering clean data samples that resemble the training distribution.

Protein discovery is both highly valuable and particularly challenging for generative models, largely due to the extreme sparsity of the protein sequence distribution \cite{romero2009exploring}. 
In this work, we focus on the task of antibody protein sequence generation. 
An antibody sequence of length $L$ is defined as a valid combination over an alphabet of 20 amino acids, 
implying a vast combinatorial search space of size $20^L$. 
Identifying valid sequences within this exponentially large space is a formidable challenge.

Existing autoregressive models struggle to generate valid protein sequences, 
suffering from error accumulation across long sequences as well as high inference latency. 
Latent diffusion models \cite{romero2009exploring}, as discrete generalizations of variational diffusion, 
alleviate some of these issues but remain memory-intensive and slow at inference \cite{frey2023protein}. 
More recently, the Discrete Walk-Jump Sampling (dWJS)\cite{frey2023protein} method, grounded in the theory of Neural Empirical Bayes \cite{saremi2019neural}, 
has been proposed as a memory-efficient alternative that also reduces inference latency. 
The dWJS framework is conceptually simple: it trains a score model at a single noise level ($\sigma$), 
optimized through denoising, and uses it during inference to guide samples in continuous space (the “walk”), 
before projecting them back to the discrete domain via a “jump” operation. 
While effective and efficient, the performance of dWJS is highly sensitive to the choice of $\sigma$. 
A large $\sigma$ oversmooths the data manifold, while a small $\sigma$ leads to sharp, fragmented regions, 
both of which might degrade performance. Careful tuning of this hyperparameter is therefore critical.  

Although existing models acknowledge the sparsity of protein data, 
we empirically observe that the degree of sparsity itself varies across the data space. 
Thus, a single global noise level may be insufficient to capture such heterogeneity. 
In this work, we propose a simple yet effective extension, \textbf{Data-Dependent Scaling with Discrete Walk-Jump Sampling (DDS-dWJS)}, 
which replaces the global $\sigma$ with data-dependent noise levels estimated from the training data. 
The only additional complexity introduced to dWJS is the estimation of these local noise scales, 
after which the score model is trained in the same manner but with sample-specific $\sigma$ values. Figure~\ref{fig:intro_contours} provides an overview of the energy surface around data points under different choices of $\sigma$. 
While Figure~\ref{fig:intro_contours} (a) and (b) illustrate the limitations of using a single global $\sigma$—either oversharpening or oversmoothing the landscape. Figure~\ref{fig:intro_contours} (d) demonstrates that with adaptive, data-dependent $\sigma \in \{0.2,1.5\}$, the energy surface naturally varies across regions according to local density. Here, the local density at a point $x$ is approximated by the inverse of its mean distance to the $4$ nearest neighbors. This heterogeneity enables the sampler to more effectively explore the diverse regions (varying density).

Our contributions are summarized as follows:
\begin{itemize}
    \item We empirically demonstrate that antibody protein sequences reside in regions of varying sparsity, motivating the need for data-dependent noise levels.
    \item On a synthetic toy dataset mimicking protein sequence distributions, we show that DDS-dWJS successfully discovers the true underlying distribution by generating valid samples.
    \item On real antibody protein sequences, we evaluate DDS-dWJS against standard dWJS and demonstrate that data-dependent scaling improves model performance.
\end{itemize}

\section{Background}
Neural Empirical Bayes (NEB) \cite{saremi2019neural} is an approach that combines non-parametric smoothing with empirical Bayes estimation to perform denoising in a least-squares sense. 
Let $x$ denote a clean sample. A smoothed version is obtained by adding Gaussian noise with standard deviation $\sigma$ (the noise level):  
\begin{equation}
    y = x + \mathcal{N}(0, \sigma^2).
\end{equation}
From the noise, the clean sample can then be estimated using the empirical Bayes formula
\begin{equation}
    \bar{x} = y + \sigma^2 \nabla \log f(y),
\end{equation}
where $f(y)$ is the density of the smoothed samples. 
Although the density $f(y)$ is generally unknown, its score function $\nabla \log f(y)$ can be approximated using a neural network. 

The neural network is trained via a denoising objective, which encourages the model to map noisy samples back to their clean versions. 
Formally, the optimization problem is given by
\begin{equation}
    L(\phi) \;=\; \mathbb{E}_{x \sim p(x), \, y \sim p(y|x)} \, \big\| x - \hat{x}_\phi(y) \big\|^2,
\end{equation}
where $p(x)$ is the distribution of the data, $p(y|x)= \mathcal{N}(x, \sigma^2)$, and the function, $\hat{x}_\phi(y) = y + \sigma^2 g_\phi(y)$

with $g_\phi(y)$ serving as the neural approximation of the score function $\nabla \log f(y)$.

The dWJS method adapts this framework to the task of protein–antibody sequence generation. 
In their formulation, each amino acid sequence is represented as a concatenated one-hot encoding: 
a sequence of length $297$ is mapped to a binary vector of size $297 \times 21$. 
Starting from this discrete representation, each sample is projected into a continuous space by adding Gaussian noise. 
A neural network is then trained to approximate the score function by learning to denoise these noisy embeddings. 

Once the score model is trained, inference proceeds as follows: 
a random sequence is generated, converted into its concatenated one-hot vector form, and perturbed with Gaussian noise to project it into the continuous space. 
In this space, the sample is iteratively updated along the direction of the learned score function (the “walk” step), effectively following the energy surface. 
After $T$ steps, the updated sample is projected back into the data space and discretized to yield a valid protein sequence (the “jump” step).

\section{Varying Sparsity of Antibody Data Distribution}


A well-recognized property of high-dimensional protein sequence data is its extreme sparsity.
Given a vocabulary size of $V = 21$ (20 amino acids plus a padding token) and sequence length $L = 297$,
the combinatorial space spans $V^L = 21^{297}$ possible sequences.
In practice, only a small fraction of these sequences is biologically valid, leaving the observed data confined to sparse regions of this exponentially large space.
Such sparsity poses a fundamental challenge for modeling the underlying distribution.
\begin{wrapfigure}{r}{0.45\linewidth}
  \centering
  \begin{minipage}{0.5\linewidth}
    \centering
    \includegraphics[width=\linewidth]{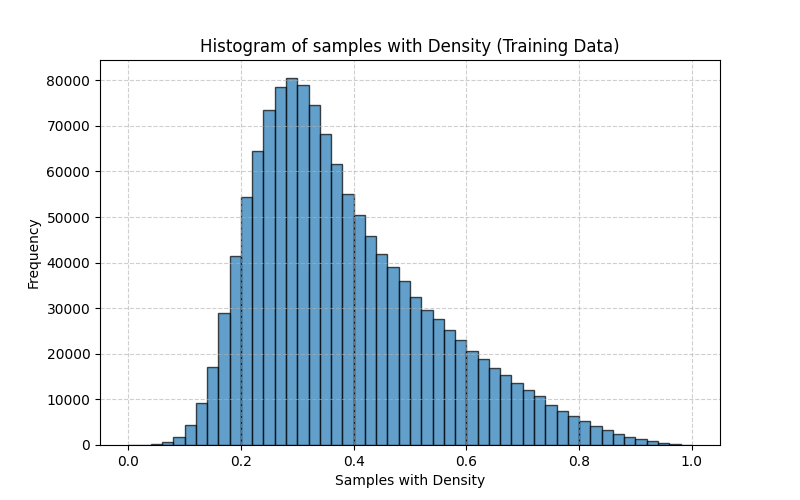}
  \end{minipage}
  \begin{minipage}{0.5\linewidth}
    \centering
    \includegraphics[width=\linewidth]{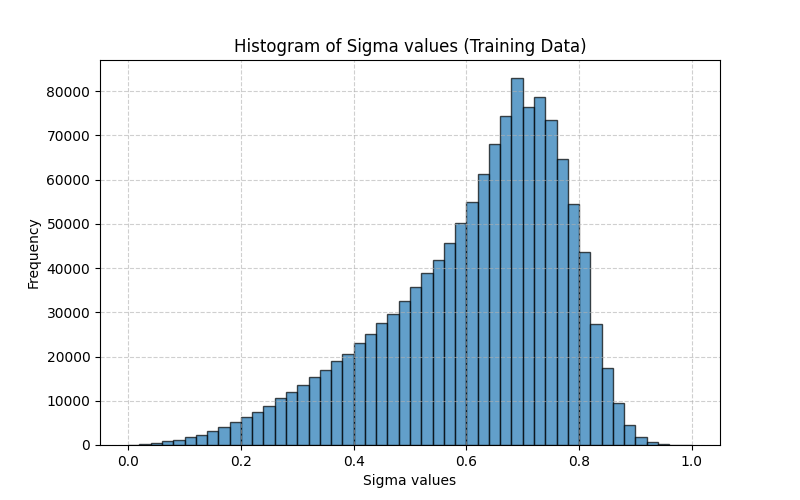}
  \end{minipage}

  \caption{Visualization of antibody dataset sparsity via KDE (left), and corresponding data-dependent noise levels $\sigma$ (right).}
  \label{fig:density_sigma}
\end{wrapfigure}

In generative frameworks such as dWJS, the choice of noise level ($\sigma$) must take this sparsity into account.
If the noise level is set too high, the data space becomes oversmoothed, leading to underfitting and loss of meaningful structure.
Conversely, if the noise level is too low, the model may overemphasize sharp local regions, making inference unstable and the sampling process harder to navigate.
Thus, careful selection of the noise level is critical for balancing fidelity and navigability in sparse, high-dimensional domains.

To analyze the sparsity of the antibody dataset, we estimated the density of training samples using a \emph{kernel density estimator (KDE)}.
KDE provides a non-parametric estimate of the probability density function by smoothing contributions from each data point.
Formally, given a dataset $\{{x_i}\}_{i=1}^N$ with feature representation $f(x_i) \in \mathbb{R}^d$, the KDE at point $f(x)$ is defined as

\begin{equation} \hat{p}\big(f(x)\big) \;=\; \frac{1}{N\,h^{d}} \sum_{i=1}^{N} K\!\left(\frac{f(x)-f(x_i)}{h}\right), \label{eq:kde} \end{equation}

where $K(\cdot)$ is a kernel function (e.g., Gaussian) and $h>0$ is the bandwidth.

For the anitbody dataset, we first transformed the high-dimensional protein sequences into six biochemical features:
(a) Hydrophobicity, (b) Molecular weight, (c) Isoelectric point, (d) Aromaticity, (e) Instability index, and (f) $\beta$-sheet content.
We applied KDE to the extracted feature vectors in order to approximate the probability density of each training sample. 
Since the dataset contains 1.3 million protein sequences, we employed Random Fourier Features \cite{rahimi2007random} to accelerate the KDE computation. 
In contrast, for the synthetic 4D toy dataset, where the number of training samples is much smaller, we constructed the KDE directly in the one-hot embedded space without approximation.

Figure~\ref{fig:density_sigma}(a) shows the normalized histogram of sample densities.
We observe that protein sequences occupy regions of widely varying density: while some sequences form dense clusters, most reside in sparse regions of the sequence space.
This highlights the heterogeneous sparsity of the POAS dataset.

Motivated by this observation, we propose to replace the conventional global noise level $\sigma$ with \emph{data-dependent} noise level $\sigma(x)$, that adapts to local density.
Specifically, we define the per-sample noise level as inversely proportional to its estimated density:
\begin{equation}
\sigma(x) \propto \frac{1}{\hat{p}\big(f(x)\big)}.
\label{eq:sigma}
\end{equation}
Thus, samples in high-density regions receive smaller $\sigma$ values, while those in sparse regions are assigned larger ones.
Figure~\ref{fig:density_sigma}(b) shows the resulting distribution of $\sigma$ values, which closely aligns with the empirical sparsity structure of the dataset.
This provides a principled mechanism to incorporate data heterogeneity into the training of our diffusion-based framework. During inference, we use $\sigma_{\max}$ both for initializing the noisy continuous samples and for the jump step during inference.


\subsection{Toy 4D Dataset Construction}

To demonstrate that our data-aware noise scaling correctly estimates the underlying distribution, 
we designed a controlled experiment on a synthetic \emph{4-dimensional toy dataset}. 
This setup is meant to mimic key aspects of protein sequences in a simplified setting.  

We consider a vocabulary of size $V=21$ (analogous to the 20 amino acids plus a padding token), 
but restrict the sequence length to $L=4$ instead of 297. 
Thus, the discrete sequence space is
\[
\mathcal{X} \;=\; \{0,1,\dots,20\}^4,
\qquad
|\mathcal{X}| \;=\; V^L \;=\; 21^4.
\]

In analogy to real protein sequences, where only a small fraction of amino-acid sequences 
yield valid proteins due to biochemical constraints, 
we impose an artificial constraint on this toy dataset.  
Specifically, each vocabulary element $v \in \{0,1,\dots,20\}$ is assigned a random 
``hydrophobicity'' score $h(v) \sim \mathrm{Uniform}[-4,4]$. 
For a candidate sequence $x = (x_1, x_2, x_3, x_4) \in \mathcal{X}$, 
we define its hydrophobicity sum as
\[
H(x) \;=\; \sum_{i=1}^4 h(x_i).
\]

A sequence is considered \emph{valid} if and only if it satisfies the threshold criterion
\[
H(x) \;\geq\; \tau,
\]
where $\tau$ is a fixed constant (e.g., $\tau=20$ in our experiments).  

Formally, the set of valid sequences is
\[
\mathcal{X}_{\text{valid}} \;=\; \{\, x \in \mathcal{X} \;:\; H(x) \geq \tau \,\}.
\]

By exhaustively enumerating all possible sequences $x \in \mathcal{X}$ 
and filtering according to the above constraint, 
we obtain a sparse distribution of valid points in 4D space. 
This mirrors the behavior of real protein sequence data, 
where the vast majority of possible sequences are invalid, 
and the observed data lie in sparse, structured regions of the combinatorial space.
\begin{figure}[t]
  \centering
  \begin{subfigure}[b]{0.33\linewidth}
    \centering
    \includegraphics[width=\linewidth]{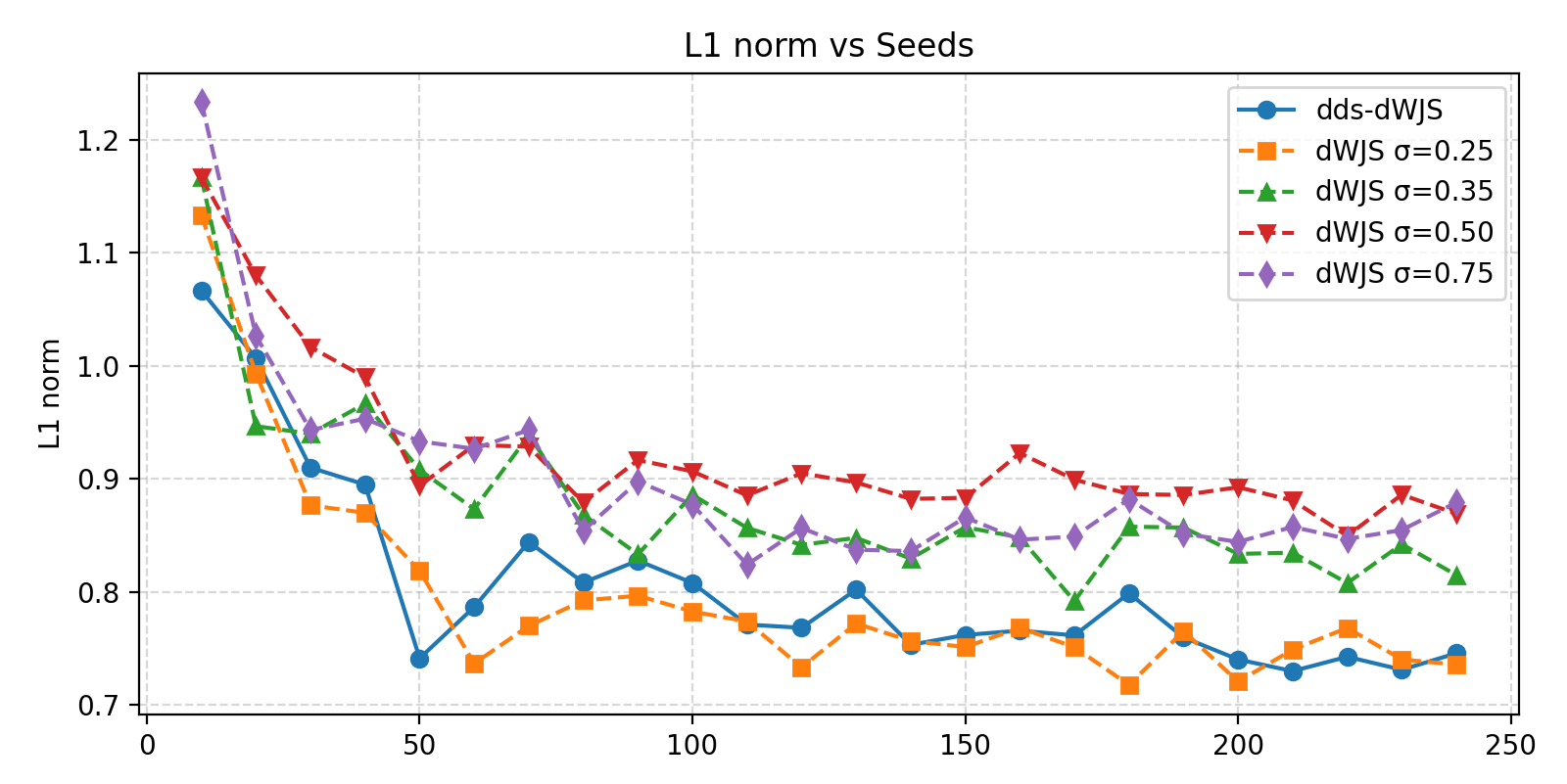}
    \caption{$L_1$ norm}
    \label{fig:l1}
  \end{subfigure}\hfill
  \begin{subfigure}[b]{0.33\linewidth}
    \centering
    \includegraphics[width=\linewidth]{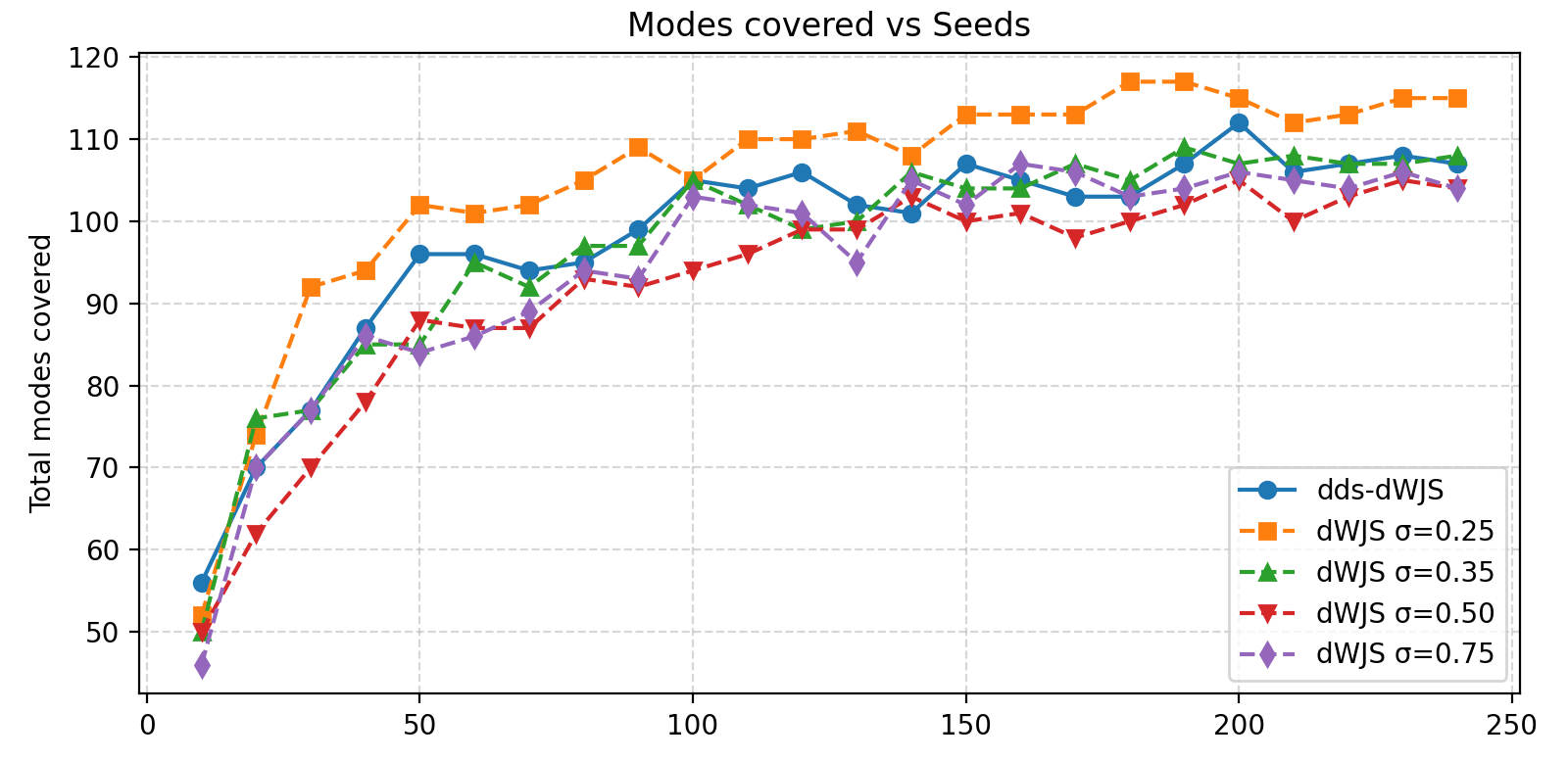}
    \caption{True modes covered}
    \label{fig:true}
  \end{subfigure}\hfill
  \begin{subfigure}[b]{0.33\linewidth}
    \centering
    \includegraphics[width=\linewidth]{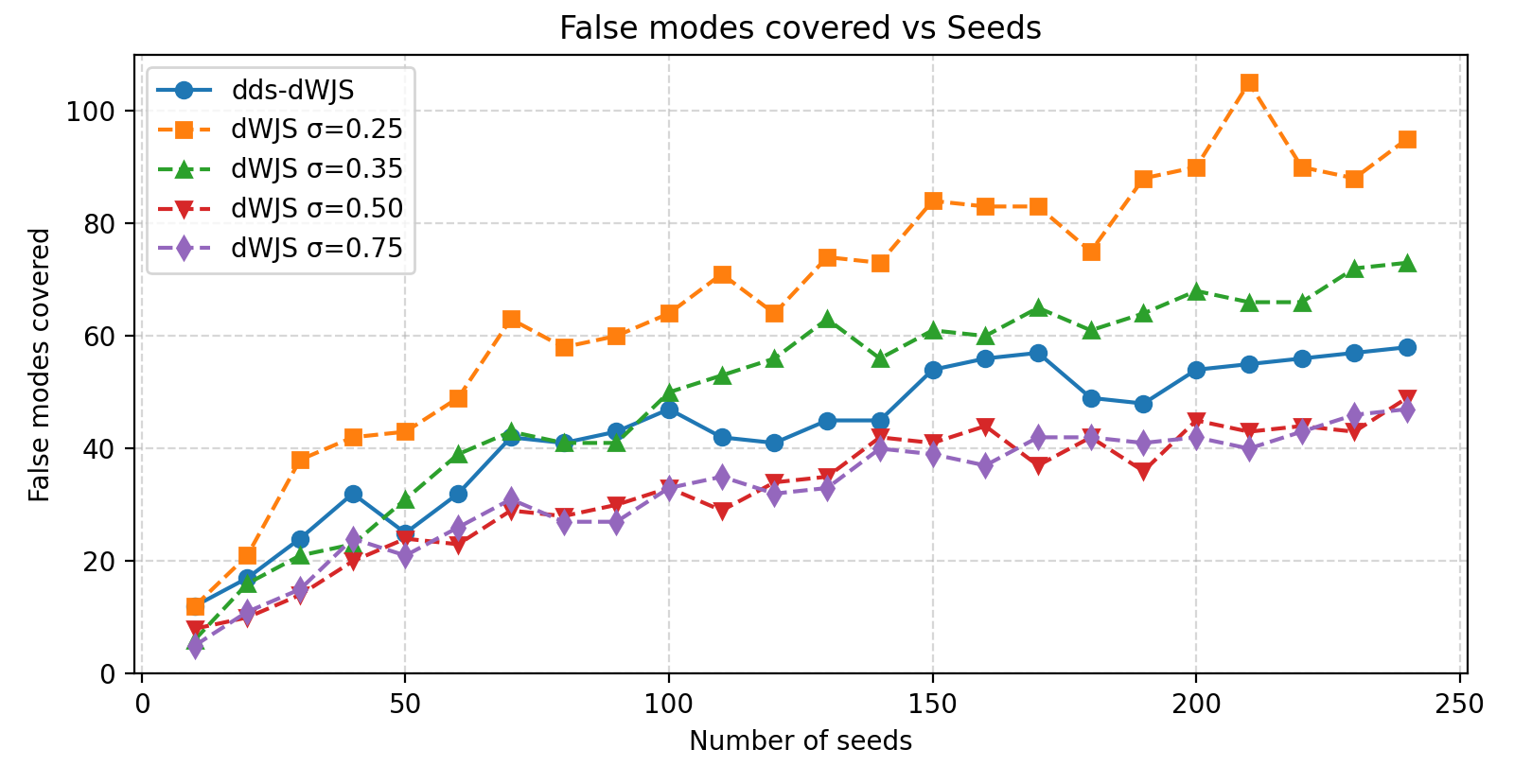}
    \caption{False modes covered}
    \label{fig:false}
  \end{subfigure}

  \caption{Performance on the synthetic 4D dataset.}
  \label{fig:synthetic_modes}
\end{figure}

We evaluate our data-dependent smoothing dWJS (DDS-dWJS) model against vanilla dWJS baselines with fixed noise scales $\sigma \in {0.25, 0.35, 0.50, 0.75}$. The synthetic dataset contains $120$ ground-truth modes, of which $50$ were used for training, and the remaining modes were held out to assess recovery. After training each model for 2000 epochs with the same ByteNet backbone, we generate samples using an increasing number of\textit{ seeds}. Total number of samples per each inference stage is $10\times$ \textit{seeds}. We evaluated the quality of generated samples on   three complementary metrics.

\paragraph{$L_1$ norm:} We measure the difference between the estimated distribution and the known true distribution. 
From Figure~\ref{fig:synthetic_modes}(a), all models show a decreasing $L_1$ norm as the number of seeds increases, 
indicating improved approximation of the underlying distribution with more generated samples. 
Among the baselines, dWJS with $\sigma=0.25$ achieves the lowest $L_1$ values overall; 
however, our DDS-dWJS closely follows its performance, providing comparable fidelity while avoiding the instability observed with larger $\sigma$ settings.
.

\paragraph{True mode recovery: } The ability of the models to recover known and discover new modes is shown in Figure~\ref{fig:synthetic_modes} (b).
Both our model and the dWJS baselines recover the majority of ground-truth modes, including held-out ones, as seeds increase. The $\sigma=0.25$ baseline shows slightly faster coverage growth in the early regime, but our model converges to similar coverage while being more stable across seeds. Larger fixed $\sigma$ values (0.5 and 0.75) recover fewer modes consistently, indicating that oversmoothing hinders exploration of the mode space.

\paragraph{False mode generation:} Figure ~\ref{fig:synthetic_modes}(c). most clearly highlights the advantage of our approach. While the $\sigma=0.25$ baseline recovers many modes, it also generates a large and steadily increasing number of false modes as seeds grow exceeding 100 by the end of the run. Our DDS-dWJS, in contrast, produces substantially fewer false modes and stabilizes at moderate levels. This suggests that density-aware noise scaling not only helps recover valid modes but also suppresses spurious generations in sparse regions of the space.
\begin{table}

\centering
\caption{Comparison of our method with baseline models across multiple metrics. 
Arrows indicate whether higher ($\uparrow$) or lower ($\downarrow$) values are better.}
\scriptsize
\begin{tabular}{lccccccc}
\toprule
\textbf{Model / Setting} & \textbf{Beta Sheet  $\uparrow$} & \textbf{Instability $\downarrow$} & $\mathbf{E_{dist}}\uparrow$  & \textbf{Intra Diversity $\uparrow$} & \textbf{FID $\downarrow$} & \textbf{DCS $\uparrow$} \\
\midrule
DDS\_dWJS(ours) & \textbf{0.3971} $\pm$ \textbf{0.0241} & \textbf{34.95} $\pm$ \textbf{6.22} & 95.80 $\pm$ 29.34  & 95.89 $\pm$ 29.20 & \textbf{0.0867} & 0.4191 $\pm$ 0.301 \\
dWJS (0.5)     & 0.3766 $\pm$ 0.0193 & 38.40 $\pm$ 6.10 & \textbf{118.11} $\pm$ \textbf{21.45} & \textbf{120.01} $\pm$ \textbf{21.10} & 4.2083 & 0.2418 $\pm$ 0.280 \\

dWJS (1.0)     & 0.3767 $\pm$ 0.0199 & 39.66 $\pm$ 6.09 & 96.27 $\pm$ 24.93  & 96.58 $\pm$ 24.58 & 0.1897 & \textbf{0.4880} $\pm$ \textbf{0.307} \\
dWJS (2.0)     & 0.3754 $\pm$ 0.0166 & 40.90 $\pm$ 5.60 & 80.46 $\pm$ 25.10 & 80.78 $\pm$ 25.04 & 0.4842 & 0.5772 $\pm$ 0.292 \\
\bottomrule
\label{tab:protein_results}
\end{tabular}
\end{table}
\subsection{Protein Antibody Sequence Generation} 

We trained our proposed DDS-dWJS model alongside the baseline dWJS using a ByteNet backbone for 40 epochs. 
For the baseline dWJS, we evaluated multiple noise-scale hyperparameters ($\sigma \in \{0.5, 1.0, 2.0\}$). 
In our DDS-dWJS model, the per-sample $\sigma$ values estimated via KDE were normalized and scaled to lie within $[0.5,1]$. 
Table~\ref{tab:protein_results} reports the results of our approach compared with the dWJS baselines. 

Overall, DDS-dWJS produces higher-quality samples, as reflected in the $\beta$-sheet content and instability indices. 
This observation is consistent with our toy dataset experiments, where DDS-dWJS was able to recover more true modes while avoiding false modes. 
Under the edit distance  ($\mathbf{E_{dist}}$) metric (comparing validation and generated sequences), dWJS($\sigma=0.5$) achieves competitive performance; 
however, the quality of sequences generated by DDS-dWJS is superior. 
A similar trend is observed for intra-diversity, where dWJS($\sigma=0.5$) performs reasonably well but fails to match the balance of diversity and quality achieved by our method. 

For the FID metric, which measures how well the generated samples cover the distribution of the training data, DDS-dWJS consistently outperforms the baselines. 
Finally, although dWJS($\sigma=0.5$) achieves the highest DCS score, prior work \cite{frey2023protein} suggests that values above $0.3$ are sufficient to generate samples that are viable in wet-lab settings. 
Thus, DDS-dWJS provides an effective balance between quality,  and distributional coverage.
\section{Conclusion}
In this work, we addressed the challenge of protein sequence generation under highly sparse and heterogeneous data distributions. 
Building on the Discrete Walk–Jump Sampling (dWJS) framework, we introduced DDS-dWJS, a data-dependent noise scaling method that assigns per-sample noise levels estimated via kernel density estimation. 
This simple yet principled extension allows the score model to adapt to local sparsity, overcoming the limitations of a single global noise scale. Through experiments on a synthetic toy dataset, we demonstrated that DDS-dWJS can faithfully recover true modes while avoiding spurious generations, thereby improving both fidelity and robustness. 
On antibody protein sequences, our method further showed consistent improvements across multiple metrics, including $\beta$-sheet content, instability index, and FID, while maintaining competitive at diversity and edit distance performance.  Overall, DDS-dWJS provides an effective approach for generative modeling in sparse, high-dimensional domains such as proteins. 
We believe that incorporating local density information into diffusion-based models opens promising directions for scalable and biologically meaningful protein design.

\section{Limitations and Future Work}

While DDS-dWJS provides a simple and effective extension to dWJS, our approach also has some limitations that motivate future research. 
Our method relies on selecting a predefined range for scaling the per-sample $\sigma$ values. The choice of lower and upper bounds can influence model performance. 
Developing a more principled or adaptive mechanism for setting these bounds remains an open problem. 

Our current study is primarily empirical. 
While the results on synthetic datasets and antibody sequences strongly suggest that data-adaptive $\sigma$ improves the quality of generated samples, a theoretical understanding of why and when adaptive noise scaling enables better approximation of the underlying score function is still lacking. 
We aim to extend this work with a formal analysis, potentially establishing error bounds that connect local density estimates, $\sigma$ selection, and the accuracy of score approximation.

{
\small
\bibliographystyle{plain}
\bibliography{main}
}

\end{document}